\documentclass[runningheads]{llncs}
\usepackage[T1]{fontenc}
\usepackage{booktabs}
\usepackage{amsmath}
\usepackage{marvosym}
\usepackage{graphicx}
\usepackage{xcolor}
 \usepackage{multirow}
\usepackage{amssymb}
\usepackage[table]{xcolor}
\usepackage[percent]{overpic}
\usepackage{hyperref}
\begin{document}
\title{SECURE: Stable Early Collision Understanding via Robust Embeddings in Autonomous Driving}
%
%
%
%
\author{
Wenjing Wang\inst{1}\textsuperscript{(\Letter)} \and
Wenxuan Wang\inst{2} \and
Songning Lai\inst{3}
}

\institute{
Xiamen University Malaysia \and
Xinjiang University \and
HKUST(GZ) \\
\email{DMT2309240@xmu.edu.my}
}
\maketitle              
\begin{abstract}
While deep learning has significantly advanced accident anticipation, the robustness of these safety-critical systems against real-world perturbations remains a major challenge. We reveal that state-of-the-art models like CRASH, despite their high performance, exhibit significant instability in predictions and latent representations when faced with minor input perturbations, posing serious reliability risks. To address this, we introduce \underline{\textbf{SECURE}} -- \underline{\textbf{S}}table \underline{\textbf{E}}arly \underline{\textbf{C}}ollision \underline{\textbf{U}}nderstanding \underline{\textbf{R}}obust \underline{\textbf{E}}mbeddings, a framework that formally defines and enforces model robustness. SECURE is founded on four key attributes: consistency and stability in both prediction space and latent feature space. We propose a principled training methodology that fine-tunes a baseline model using a multi-objective loss, which minimizes divergence from a reference model and penalizes sensitivity to adversarial perturbations. Experiments on DAD and CCD datasets demonstrate that our approach not only significantly enhances robustness against various perturbations but also improves performance on clean data, achieving new state-of-the-art results.

\keywords{Accident Anticipation, Adversarial Robustness, Autonomous Driving, Robust Embeddings, Spatio-temporal Learning }
\end{abstract}
\section{Introduction}
\label{sec:intro}

The rapid advancement of deep learning has profoundly reshaped modern intelligent transportation systems (ITS) \cite{fan2021mvit,liao2024gpt4grounding}, enabling vehicles and infrastructure to perceive complex environments, anticipate risks, and execute timely decisions. As ITS progresses toward higher levels of automation, robustness and safety become foundational requirements rather than optional enhancements \cite{li2023robustbayes,han2022backdoor}. Among the diverse modules supporting intelligent transportation, accident anticipation stands as one of the most safety-critical tasks: predicting the evolution of risk seconds before a hazardous event occurs requires not only accuracy but also robustness under real-world disturbances \cite{chan2017anticipating,suzuki2018adalea,karim2022dsta,thakur2024graphgraph}.
\begin{figure}[t]
    \centering
    \includegraphics[width=0.95\columnwidth]{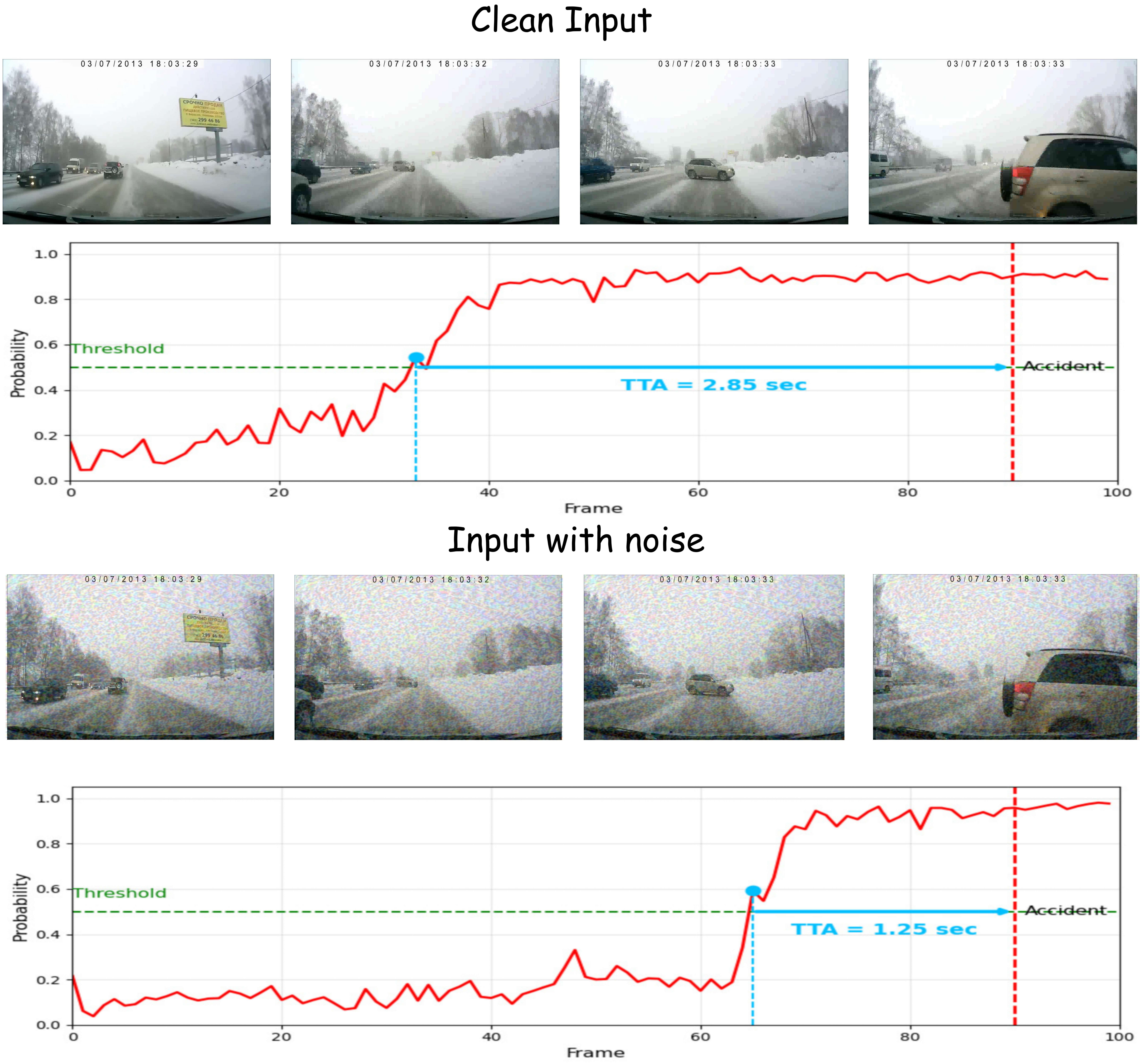}
    \caption{The CRASH model exhibits significant output differences between clean and perturbed inputs.}
    \label{fig:my_image}
\end{figure}
However, achieving reliable accident anticipation remains far from solved \cite{bao2020uncertainty,bao2021drive,karim2023multistream}. Beyond traditional prediction accuracy, modern ITS must ensure stable and robust output behaviors under a wide range of environmental changes. In high-stakes scenarios, even small perturbations—such as illumination shifts, sensor noise, temporal jitter, or sampling irregularities—can lead to significant fluctuations in model predictions \cite{madry2018,han2022backdoor}. Such instability directly threatens system reliability, complicates safety verification, and undermines dependable operation in unconstrained environments \cite{li2023robustbayes}.

In recent years, Crash Recognition and Anticipation System Harnessing (CRASH), which integrates context-aware and temporal focus mechanisms, has become a powerful representative model within ITS-related video-based risk forecasting \cite{liao2024crash}. CRASH achieves SOTA performance on benchmark datasets and can effectively recognize hazardous patterns early in temporal sequences \cite{liao2024crash}.

Although CRASH demonstrates robust predictive performance under standard benchmark settings, our research reveals a limitation that has not received sufficient attention: CRASH exhibits significant instability in both its predictive outputs and its underlying latent representations. Even minor perturbations to the input—such as sensor noise or small frame-level corruptions—can distort the model’s representation space, leading to substantial shifts in its predicted probability trajectories (see Figure \ref{fig:my_image}). This instability may result in delayed or unreliable time-to-accident estimates, posing critical safety concerns for intelligent transportation applications.

To address these challenges, we introduce the concept of \underline{\textbf{SECURE}} -- \underline{\textbf{S}}table \underline{\textbf{E}}arly \underline{\textbf{C}}ollision \underline{\textbf{U}}nderstanding \underline{\textbf{R}}obust \underline{\textbf{E}}mbeddings
and propose a robustness analysis framework for accident anticipation scenarios. This framework aims to systematically evaluate and improve the reliability of CRASH models under noise perturbations. It defines four key properties to construct the model's robustness.

Furthermore, it introduces multi-dimensional robustness attributes to define the consistency and stability of model predictions. Based on this, we conducted a comprehensive robustness analysis experiment on CRASH and SECURE. The experiment revealed the strong robustness of our proposed framework, providing a scientific basis for subsequent model optimization. Our contributions are
as follows:

\noindent(1) \textbf{Comprehensive Robustness Assessment:} We performed an in-depth robustness analysis of the CRASH model, discovering reliability flaws in the CRASH framework that make the prediction results highly unstable to input and parameter perturbations.

\noindent(2) \textbf{Rigorous Definition of SECURE:} We propose a rigorous definition of SECURE, which systematically defines what constitutes a reliable and robust accident anticipation framework. This definition can serve as a principle for evaluating and improving the robustness of accident anticipation models.

\noindent(3) \textbf{A Robust Framework for Constructing SECURE:} We propose a robustness framework that improves model robustness while maintaining excellent output by fine-tuning the parameters of the original model during training. We validate the effectiveness of this framework through experiments.

\section{Related Work}
\label{sec:related work}








The key to predicting traffic accidents lies in whether the model can extract
spatio-temporal features from complex and dynamic traffic scenarios~\cite{li2023robustbayes}.
Early approaches primarily relied on convolutional neural networks (CNNs) to
extract visual features from dashcam footage~\cite{chan2017anticipating,fang2022consistency,ma2022openworld}, typically combined with sequential
networks such as recurrent neural networks (RNNs) and long short-term memory
(LSTM) networks to preserve temporal coherence~\cite{fatima2021gfa,takimoto2019eventrecorder,ye2019anopcn}. To capture interactions between traffic
entities, recent research has introduced graph neural networks (GNNs)~\cite{karim2023multistream,thakur2024graphgraph,wang2023gsc} and Transformer-based
architectures~\cite{fan2021mvit,feng2021mist,li2022uniformerv2}, which effectively model the spatial dynamics of road users.

Recognising that not all visual regions contribute equally to accident risk,
attention mechanisms have become standard practice. Chan et al.~\cite{chan2017anticipating} introduced dynamic soft attention to fuse object-level and
frame-level features, while Karim et al.~\cite{karim2022dsta,karim2023multistream} developed spatio-temporal attention networks to prioritise hazardous
regions; Zeng et al.~\cite{zeng2017agentcentric} used attention RNNs to localise potential danger zones. Researchers have also studied interpretability:
Monjuru et al.~\cite{karim2021xai} embedded Grad-CAM into a GRU to generate semantic feature maps, and Bao et al.~\cite{bao2021drive,yao2022dota} explored
generative models and visualisations to enhance model transparency.

Despite the remarkable accuracy achieved on benchmark datasets, robustness
within open-world intelligent transport systems (ITS) remains a critical yet
understudied challenge. The random nature of traffic environments requires
models to handle not only standard dynamic variations but also extreme
outliers, sensor noise, and adversarial interference. Ma et al.~\cite{ma2022openworld} highlight the difficulty of autonomous-driving perception when
encountering unknown categories or rare events. The vast variability in
traffic scenes~\cite{li2023robustbayes} gives rise to long-tail distribution issues, where models trained on normal data fail to generalise to rare
anomalies~\cite{thakare2023rares}. Whilst some research models uncertainty---for example, Bao et al.~\cite{bao2020uncertainty} proposed a perceptual
uncertainty graph---most methods still assume relatively clean data
distributions and lack rigorous stress testing against environmental
perturbations.

Furthermore, the safety-critical nature of accident prediction demands
robust interference resistance. Han et al.~\cite{han2022backdoor} demonstrated physical backdoor attacks on lane detection systems, showing that subtle
physical perturbations can mislead deep perception models. However, most
accident prediction frameworks focus on interactions between dynamic
objects~\cite{karim2022dsta,liao2024gpt4grounding} or optimising loss functions for early detection~\cite{suzuki2018adalea}, often neglecting robustness against
visual noise, adversarial attacks, and domain shifts. Consequently, there is still no comprehensive framework that systematically benchmarks robustness and guides mitigation for accident prediction models under noise, attacks, and domain shifts.
To address these limitations, our work aims to provide a unified framework for evaluating and enhancing robustness in ITS accident prediction. By integrating insights from prior work and leveraging advanced techniques, the framework improves prediction stability and fidelity under perturbations.

\section{Method}
\label{sec:method}

\subsection{Preliminaries-CRASH}
\label{sec:CRASH}
CRASH is an accident anticipation model that predicts accident probabilities from dashboard videos. Given an input video $V = \{V_1,\dots,V_T\}$, it outputs a frame-wise probability sequence $P = \{p_1,\dots,p_T\}$, where $p_t \in [0,1]$ denotes the accident probability at frame $t$.

\noindent\textbf{Object and Context Features.} 
A Cascade R-CNN is used to detect the top-$n$ dynamic traffic agents (e.g., vehicles, pedestrians, cyclists). Their cropped regions are embedded by a VGG-16 backbone into object features $F_o \in \mathbb{R}^{n\times d}$. In parallel, a feature extractor encodes the full frame into global context features $F_c \in \mathbb{R}^{d}$, capturing lane markings, road layout, traffic signs, and other scene-level cues outside bounding boxes.

\noindent\textbf{Object-aware and Context-aware Modules.}
The object-aware module enhances the detected object features $F_o$ through an
Object Focus Attention (OFA) mechanism, producing refined object-aware features
$\tilde{F}_o$. In parallel, the context-aware module transforms the frame-level
context features $F_c$ into enhanced context-aware features $\tilde{F}_c$
through a lightweight feature refinement network.

\noindent\textbf{Temporal Fusion and Prediction.}
Temporal features are obtained by concatenating the outputs of the previous
modules and encoding them with a dual-layer GRU, producing the temporal hidden
state $h_t$. An MLP then maps $h_t$ to the accident probability $p_t$.

\noindent\textbf{Training.}
The loss function combines an anticipation loss $\mathcal{L}_a$ and an enhancement loss $\mathcal{L}_e$.
The anticipation loss encourages early prediction by applying a time-dependent penalty 
$e^{-\frac{1}{2}\max\!\left(\frac{\tau - t}{f},\,0\right)}$
to positive frames. It is defined as:


\begin{equation}
\mathcal{L}_a = \frac{1}{B}\sum_{v=1}^{B}
\Big[-l_v\sum_{t=1}^{T} e^{-\frac{1}{2}\max\big(\frac{\tau-t}{f},0\big)}\log(p_t)
-(1-l_v)\sum_{t=1}^{T}\log(1-p_t)\Big].
\end{equation}

\noindent
where, $B$ is the batch size, $l_v \in \{0,1\}$ represents the binary label of accident occurrence within each video ($1$ for an accident, $0$ for none), $T$ is the total number of frames per video, $\tau$ is the ground-truth accident time, $f$ is the frames per second (fps) of the video, and $p_t$ is the predicted accident probability at frame $t$.

To stabilize temporal representations, the enhancement loss
$\mathcal{L}_e$ supervises an auxiliary prediction $p_e$.

\begin{equation}
p_e = \phi_{\mathrm{MLP}}\!\left(
    \phi_{\mathrm{MHA}}\!\left(
        \phi_{\mathrm{PE}}(h^{2}_{1},\, h^{2}_{2},\, \ldots,\, h^{2}_{T})
    \right)
\right)
\end{equation}

\noindent
where $\phi_{\mathrm{PE}}$ denotes the position encoding mechanism, $\phi_{\mathrm{MHA}}$ denotes the multi-head self-attention mechanism, and $h_t^2$ is the hidden state of the second-layer GRU at frame $t$.

It is derived from position-encoded second-layer GRU hidden states via a multi-head self-attention block and an MLP. The enhancement loss is given by:
\begin{equation}
\mathcal{L}_e = \frac{1}{B} \sum_{v=1}^{B} \left[ -\, l_v \log(p_e) - (1 - l_v)\log(1 - p_e) \right].
\end{equation}

Finally, the overall loss function is the sum of $\mathcal{L}_a$ and $\mathcal{L}_e$ 
and incorporates homoscedastic uncertainty weighting:
\begin{equation}
\mathcal{L}_{Task} = \frac{\mu_1}{2\rho_1^{2}}\, \mathcal{L}_a 
  + \frac{\mu_2}{2\rho_2^{2}}\, \mathcal{L}_e 
  + \log(\rho_1 \rho_2),
\end{equation}

\noindent
where $\mu_1, \mu_2$ are hyperparameters and $\rho_1, \rho_2$ are learnable uncertainty coefficients initialized to $1$. This formulation allows the model to jointly optimize early accident anticipation and auxiliary temporal representation quality while adaptively balancing the two loss terms.

\subsection{Robustness Issues}




The CRASH model demonstrates state-of-the-art performance on video-based accident anticipation tasks. While CRASH exhibits strong predictive power and competitive accident prediction performance, its high sensitivity to input perturbations and parameter variations exposes significant reliability deficiencies. These issues manifest as unstable probability outputs, inconsistent attention to key subjects, and fluctuations in accuracy and estimates, even with slight changes in the input sequence.

This sensitivity undermines CRASH's reliability in real-world autonomous driving scenarios, where sensor noise, camera jitter, and environmental changes are unavoidable. Consequently, the model's decisions may deviate from underlying visual evidence, reducing interpretability and credibility in safety-critical applications. Through our research on CRASH, we propose a general solution to improve its reliability and ultimately provide general insights for enhancing the reliability and robustness of accident prediction models across the entire video understanding domain.

\subsection{The Concept of "SECURE"}

In the context of high-stakes accident anticipation, a 
\textit{SECURE Framework} refers to the capability of a model 
to consistently adhere to an optimal decision manifold while 
maintaining intrinsic invariance under perturbations. This 
notion emphasizes not only the precision of risk assessment 
but also the model’s resilience to sensory noise and the 
semantic alignment of its latent representations. By ensuring 
robustness, we aim to stabilize the anticipation process, 
making the risk evolution robust regardless of environmental 
fluctuations.

To rigorously address the robustness issues in the accident anticipation, we propose that a \textit{SECURE Framework} must encompass four fundamental attributes, structured across the Output Space and the Latent Space:

\noindent \underline{(i)} \textbf{Consistency in Prediction Space (Cps).} 
This attribute pertains to the alignment between the model's risk estimation and the theoretical optimal reference. Let $f(x; \theta)$ denote the trainable model's prediction for input $x$, and $f^*(x)$ denote the frozen model prediction. To satisfy this attribute, the divergence denoted as $D_{out}(f(x; \theta), f^*(x))$ must be bounded by a constant $\gamma_1$. This ensures that SECURE preserves the high-fidelity decision capability of the CRASH model.

\noindent\underline{(ii)} \textbf{Stability in Prediction Dynamics (Spd).}
This attribute addresses the robustness of the prediction against input perturbations without relying on external anchors. It requires SECURE to exhibit intrinsic self-consistency. Specifically, for an input $x$ and its perturbed counterpart $x+\delta$, the forecasts must converge. The difference, measured by a metric $D_{out}$, must be bounded by $\gamma_2$.

\noindent\underline{(iii)} \textbf{Consistency in Latent Manifold (Clm).}
Robustness must permeate the deep representation level. This attribute requires that the high-dimensional features extracted by SECURE, $h(x; \theta)$, semantically align with the CRASH's feature manifold $h^*(x)$. The divergence $D_{feat}(h(x; \theta), h^*(x))$ must not exceed a threshold $\beta_1$. This ensures that the model captures the correct semantic cues (e.g., motion patterns) rather than fitting noise.

\noindent\underline{(iv)} \textbf{Stability in Latent Dynamics (Sld).}
This attribute ensures that the internal feature extraction remains invariant to noise. The latent representation $h(x; \theta)$ and the perturbed representation $h(x+\delta; \theta)$ must remain within a compact neighborhood in the latent space. This prevents the "butterfly effect" where minor input noise amplifies into divergent semantic features.

\begin{figure*}[t]
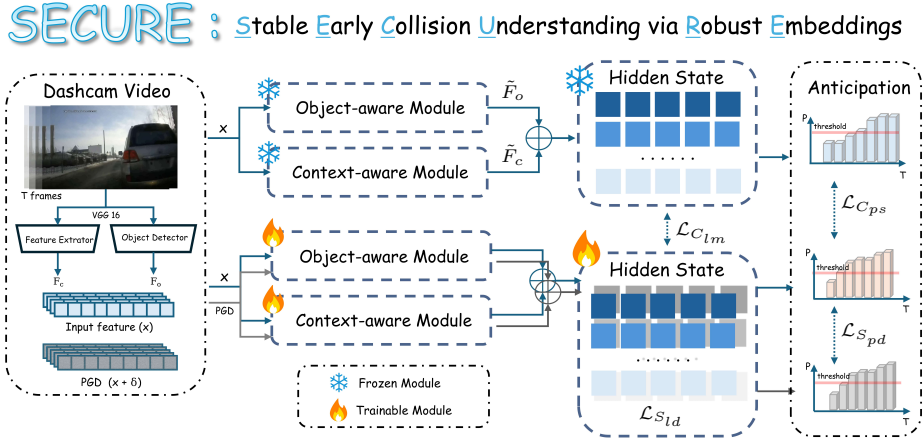

\centering
\begin{overpic}[width=\textwidth]{images/framework_final.pdf}
    \put(69,  2){\scriptsize $\mathcal{L}_{S_{ld}}$}
    \put(54, 37){\scriptsize $\tilde{F}_o$}
    \put(54, 30){\scriptsize $\tilde{F}_c$}
    \put(73, 22){\scriptsize $\mathcal{L}_{C_{lm}}$}
    \put(91, 25){\scriptsize $\mathcal{L}_{C_{ps}}$}
    \put(91, 11){\scriptsize $\mathcal{L}_{S_{pd}}$}
\end{overpic}

\caption{Overall framework of SECURE.}
\label{fig:secure_framework}
\end{figure*}

\subsection{Definition of SECURE}
\label{Definition}

Based on the attributes above, we provide a formal definition. 
A CRASH model is considered $(\gamma_1, \gamma_2, \beta_1, \beta_2, \epsilon)$-SECURE if it satisfies the following four criteria for any input features sequence $x = \{x_1,\dots,x_T\} \in \mathcal{X}$, where $\theta$ denotes the parameters of SECURE:

(i) Consistency in Prediction Space ($C_{ps}$):
\begin{equation}
D_{out}(f(x; \theta), f^*(x)) \le \gamma_1
\end{equation}

(ii) Stability in Prediction Dynamics ($S_{pd}$):
\begin{equation}
D_{out}(f(x; \theta), f(x+\delta; \theta)) \le \gamma_2,
\quad \forall \|\delta\| \le \epsilon
\end{equation}

(iii) Consistency in Latent Manifold ($C_{lm}$):
\begin{equation}
D_{feat}(h(x; \theta), h^*(x)) \le \beta_1
\end{equation}

(iv) Stability in Latent Dynamics ($S_{ld}$):
\begin{equation}
D_{feat}(h(x; \theta), h(x+\delta; \theta)) \le \beta_2,
\quad \forall \|\delta\| \le \epsilon
\end{equation}

\noindent where $D_{out}$ and $D_{feat}$ represent distance metrics 
(Mean Squared Error), $\|\cdot\|$ denotes 
a norm constraint on the perturbation $\delta$, and $\epsilon \ge 0$ 
defines the robustness radius.

\subsection{Constructing a SECURE framework}

In order to construct a $(\gamma_1, \gamma_2, \beta_1, \beta_2, \epsilon)$-robust accident anticipation model from CRASH to SECURE, we formulate a multi-objective optimization problem that aligns with the attributes defined in Section \ref{Definition} and shown in Figure \ref{fig:secure_framework}. This formulation leads to a comprehensive objective function that balances the four critical aspects of the model: Consistency in Prediction Space ($C_{ps}$), Stability in Prediction Dynamics ($S_{pd}$), Consistency in Latent Manifold ($C_{lm}$), and Stability in Latent Dynamics ($S_{ld}$).

Given the definitions of $C_{ps}$, $S_{pd}$, $C_{lm}$, and $S_{ld}$, we aim to find the optimal model parameters $\theta$ that satisfy the following optimization problem. Formally,
\begin{scriptsize}
\begin{align}
\min_{\theta}\big[
\mathcal{L}_{C_{ps}}(f(\cdot))
+ \mathcal{L}_{S_{pd}}(f(\cdot))
+ \mathcal{L}_{C_{lm}}(h(\cdot))
+ \mathcal{L}_{S_{ld}}(h(\cdot))
\big].
\end{align}
\end{scriptsize}

Each loss term corresponds to one of the four attributes of the SECURE framework:

\begin{align}
\mathcal{L}_{C_{ps}}(f(\cdot))
&= \mathbb{E}_{x}\!\left[
    D_{\text{out}}\!\left(f(x;\theta),\, f^{*}(x)\right)
\right], \\
\mathcal{L}_{S_{pd}}(f(\cdot))
&= \mathbb{E}_{x,\delta}\!\left[
    D_{\text{out}}\!\left(f(x;\theta),\, f(x+\delta;\theta)\right)
\right], \\
\mathcal{L}_{C_{lm}}(h(\cdot))
&= \mathbb{E}_{x}\!\left[
    D_{\text{feat}}\!\left(h(x;\theta),\, h^{*}(x)\right)
\right], \\
\mathcal{L}_{S_{ld}}(h(\cdot))
&= \mathbb{E}_{x,\delta}\!\left[
    D_{\text{feat}}\!\left(h(x;\theta),\, h(x+\delta;\theta)\right)
\right].
\end{align}

\noindent
where, $D_{out}$ and $D_{feat}$ are measures of divergence in the output and feature spaces, respectively. The perturbation $\delta$ is constrained by $\|\delta\| \le \epsilon$, and $\mathbb{E}_{x, \delta}$ denotes the expectation over both the input space and the perturbation space. $f^*(x)$ and $h^*(x)$ represent the mappings of the frozen CRASH model, while $f(x; \theta)$ and $h(x; \theta)$ denote the predicted output and hidden state representations of SECURE. To achieve this, we incorporate the following loss terms into the optimization process:

\begin{equation}
\mathcal{L}_{\text{Total}}
= \mathcal{L}_{\text{Task}}
+ \lambda_{c}^{out} \mathcal{L}_{C_{ps}}
+ \lambda_{s}^{out} \mathcal{L}_{S_{pd}}
+ \lambda_{c}^{feat} \mathcal{L}_{C_{lm}}
+ \lambda_{s}^{feat} \mathcal{L}_{S_{ld}}.
\label{eq:total_loss}
\end{equation}

\noindent
where, $\mathcal{L}_{\text{Task}}$ is the base loss function of CRASH, and the $\lambda$ terms are regularization coefficients that control the influence of each auxiliary robustness term.

Inspired by the Projected Gradient Descent (PGD) method described by Madry et al.~\cite{madry2018},
we iteratively update the perturbation $\delta$ and the model parameters $\theta$ such that the model 
remains robust against adversarial perturbations while maintaining consistency in both prediction 
and latent spaces. The iterative process for the $p$-th update is defined as follows.
\begin{equation}
\delta_{p} = \delta_{p-1}^*
+ \frac{\alpha_p}{|\mathcal{B}_{p-1}|}
\sum_{x \in \mathcal{B}_{p-1}}
\nabla_{\delta_{p-1}^*}
\left(
\mathcal{L}_{C_{ps}}
+ \mathcal{L}_{S_{pd}}
+ \mathcal{L}_{C_{lm}}
+ \mathcal{L}_{S_{ld}}
\right),
\end{equation}

\noindent where 
\begin{equation}
\delta_{p}^* = 
\arg\min_{\|\delta\|\le\epsilon}
\|\delta - \delta_{p}\|
\end{equation}
projects the perturbation back into the $\epsilon$-ball to satisfy the norm constraint. 
Here, $\mathcal{B}_{p-1}$ denotes the batch of samples at iteration $p-1$, and $\alpha_p$ is 
the PGD step size.

After obtaining the worst-case perturbation $\delta^*$ after $P$ iterations, the model parameters 
are updated from $\theta^{t-1}$ to $\theta^{t}$ using batched gradients derived from 
the total objective $\mathcal{L}_{\text{Total}}$. This ensures that the model adheres to the 
principles of consistency of prediction space, stability of prediction dynamics, consistency of latent manifold, and stability of latent dynamics, thereby yielding dependable accident anticipation across varying conditions.

\section{Experiment}
\label{sec:exp}
\subsection{Experimental Setup}

\noindent\textbf{Datasets and Baselines.}
We evaluate our framework on two benchmark datasets: 
the Dashcam Accident Dataset (DAD) ~\cite{chan2017anticipating} and the Car Crash Dataset (CCD)~\cite{bao2020uncertainty}, following their standard training/testing splits. We compare our proposed SECURE 
against the state-of-the-art accident anticipation baseline 
CRASH to validate the effectiveness of our consistency 
and stability constraints.

\medskip
\noindent\textbf{Implementation Details.}
We employ Projected Gradient Descent (PGD) attacks. The attack hyperparameters are configured as follows: perturbation budget 
$\epsilon = 0.01$, step size $\alpha = 0.002$, and 
$P=20$ attack iterations. All experiments are conducted on an NVIDIA GeForce RTX~4090 GPU. 
We train the model using the Adam optimizer with an initial 
learning rate of $1 \times 10^{-4}$ and a batch size of 10. 
The hyperparameters for the robustness loss terms are set to: 
output stability $\lambda_{s}^{\text{out}} = 50.0$, output 
consistency $\lambda_{c}^{\text{out}} = 50.0$, and 
feature-level $ \lambda_{s}^{feat}
=\lambda_{c}^{feat} = 0.01$.

\subsection{Performance Results}

Table~\ref{tab:baseline_results} presents the test results of our proposed 
SECURE framework and the baseline method CRASH under conditions without 
added noise perturbations. Following the evaluation protocol in CRASH, we 
adopt Average Precision (AP) and Mean Time-to-Accident (mTTA) as the 
performance metrics. Specifically, AP serves as a comprehensive measure of 
model accuracy, reflecting prediction consistency across varying decision 
thresholds by computing the area under the precision--recall curve. 
Meanwhile, mTTA evaluates the timeliness of the model’s predictions by 
measuring the average time interval between the model's first risk 
prediction (exceeding a predefined threshold) and the actual time of 
an accident. A higher mTTA value indicates that the model can anticipate 
accidents earlier, providing drivers with more reaction time.

As shown in Table~\ref{tab:baseline_results}, our framework SECURE outperforms CRASH and 
achieves SOTA performance on clean data without any added 
perturbations. We infer that this improvement stems from the inherent noise 
contained in real-world driving data. By mitigating the influence of such 
noise during training, our framework enhances model prediction quality. 
Importantly, SECURE does not sacrifice original performance in exchange for 
robustness; instead, it achieves even better results while maintaining 
stronger stability.

\begin{table}[t]
\centering
\caption{Comparison of model performance on the complete datasets. Bold and underlined
values represent the best and second-best performance in
each category.}
\begin{tabular}{lcccc}
\toprule
\textbf{Model} 
& \multicolumn{2}{c}{\textbf{DAD}~\cite{chan2017anticipating}}
& \multicolumn{2}{c}{\textbf{CCD}~\cite{bao2020uncertainty}} \\
\cmidrule(lr){2-3}
\cmidrule(lr){4-5}
& AP(\%)$\uparrow$ & mTTA(s)$\uparrow$
& AP(\%)$\uparrow$ & mTTA(s)$\uparrow$ \\
\midrule
DSA~\cite{chan2017anticipating}         & 48.1 & 1.34 & \underline{99.6} & 4.53 \\
L-RAI~\cite{zeng2017agentcentric}      & 51.4 & 3.01 & 98.9 & 3.32 \\
AdaLEA~\cite{suzuki2018adalea}   & 62.3 & 2.94 & 99.2 & 3.45 \\
DSTA~\cite{karim2022dsta}      & 59.2 & 2.60 & 98.8  & 4.53 \\
UString~\cite{bao2020uncertainty} & 58.0 & 2.85 & 99.5 & 4.52 \\
GSC~\cite{wang2023gsc}         & 60.4 & 2.55 & 99.3 & 3.58 \\
CRASH~\cite{liao2024crash}     & \underline{67.2} & \underline{3.02} & 99.5 & \underline{4.55} \\
\rowcolor{gray!30}
\textbf{SECURE}        & \textbf{68.7} & \textbf{3.13} & \textbf{99.7} & \textbf{4.58} \\
\bottomrule
\end{tabular}
\label{tab:baseline_results}
\end{table}

\begin{table*}[t!]
  \centering
  \caption{Comparison of CRASH and SECURE on DAD and CCD datasets under different perturbations. The number in the parentheses is the standard deviation of Gaussian noise. }
  \label{tab:drive_dad_ccd}
  \small
  \resizebox{\textwidth}{!}{
  \begin{tabular}{lcccccccccccc}
    \toprule
    \multirow{2}{*}{\textbf{Model / Dataset}}
    & \multicolumn{2}{c}{Clean} 
    & \multicolumn{2}{c}{IP (0.1)} 
    & \multicolumn{2}{c}{IP (0.2)}
    & \multicolumn{2}{c}{LP (0.1)} 
    & \multicolumn{2}{c}{LP (0.2)}  \\
    \cmidrule(lr){2-3}
    \cmidrule(lr){4-5}
    \cmidrule(lr){6-7}
    \cmidrule(lr){8-9}
    \cmidrule(lr){10-11}
    & AP(\%)$\uparrow$ & mTTA(s)$\uparrow$
    & AP(\%)$\uparrow$ & mTTA(s)$\uparrow$
    & AP(\%)$\uparrow$ & mTTA(s)$\uparrow$
    & AP(\%)$\uparrow$ & mTTA(s)$\uparrow$
    & AP(\%)$\uparrow$ & mTTA(s)$\uparrow$ \\
    \midrule

    
    \textbf{CRASH (DAD)}
    & 67.21 & 3.020
    & 65.83 & 3.053
    & 63.37 & 2.964
    & 64.41 & 2.984
    & 44.73 & 3.119 \\

    \rowcolor{gray!30}
    \textbf{SECURE (DAD)}
    & 68.67 & 3.127
    & 68.42 & 3.120
    & 68.32 & 3.062
    & 68.19 & 3.083
    & 57.61 & 3.207\\
    
    \midrule

    \textbf{CRASH (CCD)}
    & 99.53 & 4.554
    & 99.29 & 4.540
    & 99.17 & 3.965
    & 98.88 & 3.074
    & 69.26 & 3.413\\

    \rowcolor{gray!30}
    \textbf{SECURE (CCD)}
    & 99.73 & 4.583
    & 99.67 & 4.483
    & 99.48 & 4.481
    & 99.29 & 3.191
    & 81.45 & 3.430\\

    \bottomrule
  \end{tabular}}
\end{table*}

\subsection{Robustness Results}

To evaluate robustness, we introduce two types of perturbations during testing:

\noindent\textbf{(1) Input Perturbation.}
Gaussian noise is added to each input frame-level feature, expressed as $x_t' = x_t + \mathcal{N}(0, \sigma)$.

\noindent\textbf{(2) Latent Perturbation.}
Since the second-layer GRU produces the model’s key temporal latent representation, we inject Gaussian noise into its parameters $\theta_{\mathrm{GRU2}}' = \theta_{\mathrm{GRU2}} + \mathcal{N}(0, \sigma)$. 

These perturbations are applied during evaluation to assess model stability under external input noise and internal temporal-module disturbance, as shown in Table~\ref{tab:drive_dad_ccd}. This table reports the performance of the baseline model and our framework under both input perturbation and intermediate-layer perturbation conditions.

Under input perturbation (IP) conditions, the CRASH
model’s performance decreased significantly, indicating that
input perturbations do indeed affect the precision of the
CRASH model. In the DAD dataset, the model’s output AP
decreased from 67.21 to 63.37, while in the CCD dataset,
the CRASH model’s mTTA decreased from 4.554 to
3.965. Under latent perturbation (LP) conditions, the impact
of perturbations on CRASH was more significant, especially
in terms of the AP metric. In the DAD and CCD datasets, the
model’s output AP decreased from 67.21 to 44.73 and from
99.53 to 69.26, respectively.

Notably, compared with the larger performance fluctuations of CRASH under perturbations, SECURE exhibits more stable behavior. In terms of average precision, SECURE shows smaller degradation across both input perturbation and latent perturbation settings, while achieving generally higher mTTA than CRASH. These results suggest that the PGD-based adversarial optimization used in SECURE improves robustness to both external feature noise and internal parameter disturbances, rather than trading accuracy for robustness. 

\subsection{Ablation Study}

\begin{table}[t]
\centering
\caption{Ablation Results of SECURE on the DAD Dataset under IP (0.2)}
\label{tab:ablation_loss}
\renewcommand{\arraystretch}{1.2}
\begin{tabular}{c c c c c c c c}
\toprule
$\mathcal{L}_{\text{Task}}$ 
& $\mathcal{L}_{C_{ps}}$ 
& $\mathcal{L}_{S_{pd}}$ 
& $\mathcal{L}_{C_{lm}}$
& $\mathcal{L}_{S_{ld}}$ 
& \textbf{AP(\%) $\uparrow$} 
& \textbf{mTTA(s)$\uparrow$} \\
\midrule

$\checkmark$ & $\times$ & $\times$ & $\times$ & $\times$
& 63.37 & 2.964\\

$\checkmark$ & $\checkmark$ & $\times$ & $\times$ & $\times$
& 63.47 & 3.036\\

$\checkmark$ & $\checkmark$ & $\checkmark$ & $\times$ & $\times$
& 64.80 & 3.015 \\

$\checkmark$ & $\checkmark$ & $\checkmark$ & $\checkmark$ & $\times$
& 65.29 & 3.001\\

\rowcolor{gray!30}
$\checkmark$ & $\checkmark$ & $\checkmark$ & $\checkmark$ & $\checkmark$
& \textbf{68.32} & \textbf{3.062} \\

\bottomrule
\end{tabular}
\end{table}

In our ablation study, we comprehensively evaluated each component detailed in Equation~\eqref{eq:total_loss}, aiming to assess the importance and effectiveness of each component in improving model performance. To this end, we designated the first term, $\mathcal{L}_{Task}$, as our principal loss function. We then systematically evaluated various configurations of $\mathcal{L}_{C_{ps}}$, $\mathcal{L}_{S_{pd}}$, $\mathcal{L}_{C_{lm}}$, and $\mathcal{L}_{S_{ld}}$ by selectively omitting these regularization terms.

Our results (as shown in Table~\ref{tab:ablation_loss}) clearly demonstrate that each regularization term in the objective function is crucial and effective. Each component makes a unique contribution to improving model performance.

\section{Conclusion}
\label{sec:con}

In this paper, we address the critical yet under-explored issue of robustness in accident anticipation models. We introduce SECURE, a principled framework that formalizes model reliability through four key attributes: consistency and stability in both the prediction and latent spaces. To construct a SECURE model, we propose a multi-objective adversarial optimizing scheme that minimizes sensitivity to perturbations while maintaining high fidelity. Experiments on the DAD and CCD datasets demonstrate that our approach not only significantly enhances stability against perturbations but also achieves SOTA on clean data. This work lays a crucial foundation for developing trustworthy accident prediction systems and offers a new standard for evaluating the robustness of safety-critical AI modules in autonomous driving.

\bibliographystyle{splncs04}
\bibliography{main}

@inproceedings{liao2024crash,
  author    = {Haicheng Liao and Haowei Sun and Huanming Shen and Chengyue Wang 
               and Chiyun Tian and Ka Lok Tam and Zhenning Li and others},
  title     = {CRASH: Crash Recognition and Anticipation System Harnessing 
               Context-Aware and Temporal Focus Attentions},
  booktitle = {Proceedings of the 32nd ACM International Conference on Multimedia},
  year      = {2024},
  pages     = {11041--11050},
  month     = oct
}

@inproceedings{bao2020uncertainty,
  author    = {Wentao Bao and Qi Yu and Yu Kong},
  title     = {Uncertainty-based Traffic Accident Anticipation with Spatio-Temporal Relational Learning},
  booktitle = {Proceedings of the 28th ACM International Conference on Multimedia (MM '20)},
  year      = {2020}
}

@inproceedings{bao2021drive,
  author    = {Wentao Bao and Qi Yu and Yu Kong},
  title     = {DRIVE: Deep Reinforced Accident Anticipation with Visual Explanation},
  booktitle = {Proceedings of the IEEE/CVF International Conference on Computer Vision (ICCV)},
  year      = {2021},
  pages     = {7619--7628}
}

@inproceedings{chan2017anticipating,
  author    = {Fu{-}Hsiang Chan and Yu{-}Ting Chen and Yu Xiang and Min Sun},
  title     = {Anticipating Accidents in Dashcam Videos},
  booktitle = {Computer Vision -- ACCV 2016},
  publisher = {Springer},
  year      = {2017},
  pages     = {136--153}
}

@inproceedings{fan2021mvit,
  author    = {Haoqi Fan and Bo Xiong and Karttikeya Mangalam and Yanghao Li and Zhicheng Yan and Jitendra Malik and Christoph Feichtenhofer},
  title     = {Multiscale Vision Transformers},
  booktitle = {Proceedings of the IEEE/CVF International Conference on Computer Vision (ICCV)},
  year      = {2021},
  pages     = {6824--6835}
}

@article{fang2022consistency,
  author  = {Jianwu Fang and Jiahuan Qiao and Jie Bai and Hongkai Yu and Jianru Xue},
  title   = {Traffic Accident Detection via Self-Supervised Consistency Learning in Driving Scenarios},
  journal = {IEEE Transactions on Intelligent Transportation Systems},
  volume  = {23},
  number  = {7},
  pages   = {9601--9614},
  year    = {2022}
}

@inproceedings{fatima2021gfa,
  author    = {Mishal Fatima and Muhammad Umar Karim Khan and Chong{-}Min Kyung},
  title     = {Global Feature Aggregation for Accident Anticipation},
  booktitle = {2020 25th International Conference on Pattern Recognition (ICPR)},
  year      = {2021},
  pages     = {2809--2816},
  publisher = {IEEE}
}

@inproceedings{feng2021mist,
  author    = {Jia{-}Chang Feng and Fa{-}Ting Hong and Wei{-}Shi Zheng},
  title     = {MIST: Multiple Instance Self-Training Framework for Video Anomaly Detection},
  booktitle = {Proceedings of the IEEE/CVF Conference on Computer Vision and Pattern Recognition (CVPR)},
  year      = {2021},
  pages     = {14009--14018}
}

@inproceedings{han2022backdoor,
  author    = {Xingshuo Han and Guowen Xu and Yuan Zhou and Xuehuan Yang and Jiwei Li and Tianwei Zhang},
  title     = {Physical Backdoor Attacks to Lane Detection Systems in Autonomous Driving},
  booktitle = {Proceedings of the 30th ACM International Conference on Multimedia},
  year      = {2022},
  pages     = {2957--2968}
}

@article{karim2022dsta,
  author  = {Muhammad Monjurul Karim and Yu Li and Ruwen Qin and Zhaozheng Yin},
  title   = {A Dynamic Spatial-Temporal Attention Network for Early Anticipation of Traffic Accidents},
  journal = {IEEE Transactions on Intelligent Transportation Systems},
  volume  = {23},
  number  = {7},
  pages   = {9590--9600},
  year    = {2022}
}

@article{karim2023multistream,
  author  = {Muhammad Monjurul Karim and Zhaozheng Yin and Ruwen Qin},
  title   = {An Attention-Guided Multistream Feature Fusion Network for Early Localization of Risky Traffic Agents in Driving Videos},
  journal = {IEEE Transactions on Intelligent Vehicles},
  year    = {2023}
}

@article{li2022uniformerv2,
  author  = {Kunchang Li and Yali Wang and Yinan He and Yizhuo Li and Yi Wang and Limin Wang and Yu Qiao},
  title   = {UniFormerV2: Spatiotemporal Learning by Arming Image {ViTs} with Video UniFormer},
  journal = {arXiv preprint arXiv:2211.09552},
  year    = {2022}
}

@article{li2023robustbayes,
  author  = {Zhenning Li and Haicheng Liao and Ruru Tang and Guofa Li and Yunjian Li and Chengzhong Xu},
  title   = {Mitigating the Impact of Outliers in Traffic Crash Analysis: A Robust Bayesian Regression Approach with Application to Tunnel Crash Data},
  journal = {Accident Analysis \& Prevention},
  volume  = {185},
  pages   = {107019},
  year    = {2023}
}

@article{liao2024gpt4grounding,
  author  = {Haicheng Liao and Huanming Shen and Zhenning Li and Chengyue Wang and Guofa Li and Yiming Bie and Chengzhong Xu},
  title   = {GPT-4 Enhanced Multimodal Grounding for Autonomous Driving: Leveraging Cross-Modal Attention with Large Language Models},
  journal = {Communications in Transportation Research},
  volume  = {4},
  pages   = {100116},
  year    = {2024}
}

@inproceedings{ma2022openworld,
  author    = {Zeyu Ma and Yang Yang and Guoqing Wang and Xing Xu and Heng Tao Shen and Mingxing Zhang},
  title     = {Rethinking Open-World Object Detection in Autonomous Driving Scenarios},
  booktitle = {Proceedings of the 30th ACM International Conference on Multimedia},
  year      = {2022},
  pages     = {1279--1288}
}

@article{karim2021xai,
  author  = {Muhammad Monjurul Karim and Yu Li and Ruwen Qin},
  title   = {Towards Explainable Artificial Intelligence (XAI) for Early Anticipation of Traffic Accidents},
  journal = {arXiv e-prints},
  year    = {2021},
  note    = {arXiv:2108.}
}

@inproceedings{takimoto2019eventrecorder,
  author    = {Yoshiaki Takimoto and Yusuke Tanaka and Takeshi Kurashima and Shuhei Yamamoto and Maya Okawa and Hiroyuki Toda},
  title     = {Predicting Traffic Accidents with Event Recorder Data},
  booktitle = {Proceedings of the 3rd ACM SIGSPATIAL International Workshop on Prediction of Human Mobility},
  year      = {2019},
  pages     = {11--14}
}

@article{thakare2023rares,
  author  = {Kamalakar Vijay Thakare and Debi Prosad Dogra and Heeseung Choi and Haksub Kim and Ig{-}Jae Kim},
  title   = {Rareanom: A Benchmark Video Dataset for Rare Type Anomalies},
  journal = {Pattern Recognition},
  volume  = {140},
  pages   = {109567},
  year    = {2023}
}

@inproceedings{thakur2024graphgraph,
  author    = {Nupur Thakur and PrasanthSai Gouripeddi and Baoxin Li},
  title     = {Graph(Graph): A Nested Graph-Based Framework for Early Accident Anticipation},
  booktitle = {Proceedings of the IEEE/CVF Winter Conference on Applications of Computer Vision (WACV)},
  year      = {2024},
  pages     = {7533--7541}
}

@article{wang2023gsc,
  author  = {Tianhang Wang and Kai Chen and Guang Chen and Bin Li and Zhijun Li and Zhengfa Liu and Changjun Jiang},
  title   = {GSC: A Graph and Spatio-Temporal Continuity Based Framework for Accident Anticipation},
  journal = {IEEE Transactions on Intelligent Vehicles},
  year    = {2023}
}

@article{yao2022dota,
  author  = {Yu Yao and Xizi Wang and Mingze Xu and Zelin Pu and Yuchen Wang and Ella Atkins and David Crandall},
  title   = {DoTA: Unsupervised Detection of Traffic Anomaly in Driving Videos},
  journal = {IEEE Transactions on Pattern Analysis and Machine Intelligence},
  year    = {2022}
}

@inproceedings{ye2019anopcn,
  author    = {Muchao Ye and Xiaojiang Peng and Weihao Gan and Wei Wu and Yu Qiao},
  title     = {ANOPCN: Video Anomaly Detection via Deep Predictive Coding Network},
  booktitle = {Proceedings of the 27th ACM International Conference on Multimedia},
  year      = {2019},
  pages     = {1805--1813}
}

@inproceedings{zeng2017agentcentric,
  author    = {Kuo{-}Hao Zeng and Shih{-}Han Chou and Fu{-}Hsiang Chan and Juan Carlos Niebles and Min Sun},
  title     = {Agent-Centric Risk Assessment: Accident Anticipation and Risky Region Localization},
  booktitle = {Proceedings of the IEEE Conference on Computer Vision and Pattern Recognition (CVPR)},
  year      = {2017},
}

@inproceedings{suzuki2018adalea,
  author    = {Tomoyuki Suzuki and Hirokatsu Kataoka and Yoshimitsu Aoki and Yutaka Satoh},
  title     = {Anticipating Traffic Accidents with Adaptive Loss and Large-Scale Incident {DB}},
  booktitle = {Proceedings of the IEEE Conference on Computer Vision and Pattern Recognition (CVPR)},
  year      = {2018},
  pages     = {3521--3529}
}

@inproceedings{madry2018,
  title={Towards deep learning models resistant to adversarial attacks},
  author={Madry, Aleksander and Makelov, Aleksandar and Schmidt, Ludwig and Tsipras, Dimitris and Vladu, Adrian},
  booktitle={International Conference on Learning Representations},
  year={2018}
}

\end{document}